\documentclass{article}

\usepackage{microtype}
\usepackage{graphicx}
\usepackage{subcaption}
\usepackage{stfloats}
\usepackage{booktabs}
\usepackage{hyperref}

\usepackage[accepted]{icml2026}

\usepackage{amsmath}
\usepackage{amssymb}
\usepackage{mathtools}
\usepackage{amsthm}
\usepackage{xspace}
\usepackage[capitalize,noabbrev]{cleveref}
\usepackage[skins]{tcolorbox}
\usepackage{tikz}
\usetikzlibrary{arrows.meta,positioning,shapes.geometric,fit,calc}

\newcommand{\method}{LeanFlow\xspace}
\newcommand{\kimi}{Kimi-K2.6\xspace}
\newcommand{\gpt}{GPT-5.5\xspace}
\newcommand{\lean}{Lean\xspace}
\newcommand{\mathlib}{Mathlib\xspace}

\definecolor{epfBlue}{HTML}{245A8D}
\definecolor{epfTeal}{HTML}{287D7D}
\definecolor{epfOrange}{HTML}{B55E1D}
\definecolor{epfGreen}{HTML}{3D7A3E}
\definecolor{epfGray}{HTML}{4A5568}
\definecolor{epfCodeBg}{HTML}{F7F9FB}
\definecolor{epfCodeRule}{HTML}{B8C2CC}
\definecolor{leanKeyword}{HTML}{6D4AA2}
\definecolor{leanGoal}{HTML}{247A5A}
\definecolor{feedbackFrame}{HTML}{287D7D}
\definecolor{feedbackBg}{HTML}{F0F7F7}
\definecolor{leanError}{HTML}{9E2F2F}
\definecolor{leanComment}{HTML}{6B7280}
\definecolor{leanType}{HTML}{1F6F8B}
\newtcolorbox{leanexamplebox}[1][]{
  enhanced,
  colback=epfCodeBg,
  colframe=epfCodeRule,
  boxrule=0.45pt,
  arc=1pt,
  left=1.4mm,
  right=1.4mm,
  top=0.9mm,
  bottom=0.9mm,
  boxsep=0pt,
  width=0.96\columnwidth,
  fontupper=\ttfamily\scriptsize,
  colbacktitle=epfCodeRule!20,
  coltitle=black!80,
  fonttitle=\sffamily\scriptsize\bfseries,
  boxed title style={boxrule=0pt, arc=1pt, colback=epfCodeRule!20},
  attach boxed title to top left={xshift=1.1mm,yshift=-0.35mm},
  #1
}
\newtcolorbox{leanfeedbackbox}{
  enhanced,
  colback=white,
  colframe=black!18,
  boxrule=0.3pt,
  arc=1pt,
  left=0.9mm,
  right=0.9mm,
  top=0.7mm,
  bottom=0.7mm,
  boxsep=0pt,
  width=0.92\columnwidth,
  fontupper=\ttfamily\scriptsize,
  borderline west={1.8pt}{0pt}{leanGoal!75},
  before skip=1.0mm,
  after skip=1.0mm
}
\newtcolorbox{leanerrorbox}{
  enhanced,
  colback=white,
  colframe=black!18,
  boxrule=0.3pt,
  arc=1pt,
  left=0.9mm,
  right=0.9mm,
  top=0.7mm,
  bottom=0.7mm,
  boxsep=0pt,
  width=0.92\columnwidth,
  fontupper=\ttfamily\scriptsize,
  borderline west={1.8pt}{0pt}{leanError!75},
  before skip=1.0mm,
  after skip=1.0mm
}
\tikzset{
  epfArrow/.style={-{Latex[length=2.2mm]}, thick, draw=epfGray},
  epfStage/.style={
    draw=epfBlue!80,
    fill=epfBlue!7,
    rounded corners=2pt,
    thick,
    align=center,
    minimum height=9mm,
    text width=18mm,
    inner sep=2pt
  },
  epfGate/.style={
    draw=epfOrange!85,
    fill=epfOrange!9,
    rounded corners=2pt,
    thick,
    align=center,
    minimum height=9mm,
    text width=18mm,
    inner sep=2pt
  },
  epfStore/.style={
    draw=epfGreen!75,
    fill=epfGreen!8,
    rounded corners=2pt,
    thick,
    align=center,
    minimum height=8mm,
    inner sep=3pt
  },
  epfTool/.style={
    draw=epfTeal!80,
    fill=epfTeal!8,
    rounded corners=2pt,
    thick,
    align=center,
    text width=26mm,
    minimum height=8mm,
    inner sep=3pt
  },
  epfDecision/.style={
    diamond,
    aspect=2.0,
    draw=epfOrange!85,
    fill=epfOrange!8,
    thick,
    align=center,
    inner sep=1pt,
    text width=18mm
  },
  epfGroup/.style={
    draw=#1!70,
    dashed,
    rounded corners=4pt,
    thick,
    inner xsep=3mm,
    inner ysep=3mm
  },
  epfStateLink/.style={
    -{Latex[length=1.5mm]},
    densely dotted,
    draw=epfGreen!75,
    line width=0.55pt,
    shorten >=0.5mm,
    shorten <=0.5mm
  }
}

\icmltitlerunning{\method: A case study in workflow-driven Lean agents}

\begin{document}

\twocolumn[
\icmltitle{\method: A Case Study in Workflow-Driven Lean Autoformalization}

  \begin{icmlauthorlist}
    \icmlauthor{Lazar Miliki\'c}{epfl}
    \icmlauthor{Simon Guilloud}{epfl}
    \icmlauthor{Khanh Nguyen}{epfl}
    \icmlauthor{Viktor Kun\v{c}ak}{epfl}
  \end{icmlauthorlist}

  \icmlaffiliation{epfl}{EPFL, Lausanne, Switzerland}
  \icmlcorrespondingauthor{Lazar Miliki\'{c}}{lazar.milikic@epfl.ch}
  \icmlkeywords{autoformalization, theorem proving, Lean, LLM agents, document-level formalization, proof repair}

  \vskip 0.3in
]

\printAffiliationsAndNotice{}

\begin{abstract}
We present and evaluate \method, an LLM agent system specialized for translating mathematical papers into buildable \lean projects.
Recent verifier-in-the-loop systems show that large formal artifacts can be produced, but it remains unclear which runtime mechanisms affect completion, auditability, or efficiency in document-to-project formalization.
We study this question through case studies on two previously unformalized mathematical papers in number theory and measure theory, using model, proof-workflow, and toolset ablations with \kimi and \gpt; we report task outcome, API calls, input tokens, and output tokens.
With \kimi, the full workflow completes both document-level projects within the 2000-call budget, while no-queue variants reach the budget limit; with \gpt, all document-level variants complete, and the full workflow has the lowest or tied-lowest input-token cost on both sources.
As complementary calibration, \method reaches 75.7\% BEq+ on the PFR slice of RLM25 and solves all five ICML 2026 AI for Math TCS challenge projects in our \gpt runs.
\end{abstract}

\section{Introduction}

Autoformalization is moving from isolated theorem translation toward turning whole mathematical sources into \lean projects.
Early language-model formalization work focused on translating individual natural-language statements into proof-assistant syntax~\citep{wu2022autoformalization}, and many theorem-proving benchmarks still ask the model to prove one supplied formal statement at a time~\citep{zheng2022minif2f,azerbayev2023proofnet}.
A document-level agent faces a different problem: it must convert mathematical prose, \TeX{} structure, references, definitions, and proofs into a buildable \lean project while preserving the intended statements and managing a long sequence of dependent proof obligations.
M2F~\citep{wang2026m2f} and recent textbook- or project-scale systems~\citep{gloeckle2026automatictextbookformalization,mathinc2025gauss,hariharan2026milestoneformalizationspherepacking,tsoukalas2026advancingmathematicsresearchaidriven} show that verifier-in-the-loop formalization can produce large \lean artifacts; we examine which parts of the runtime help agents keep the source claims fixed and finish the resulting proof work.

Motivated by this document-scale setting, we introduce \method as a \lean-specific runtime that separates mathematical editing from workflow control.
It first runs a deterministic source preflight: the runtime resolves the input document, extracts or records source blocks, labels, references, PDFs, figures, and support files, and refuses ambiguous \TeX{} roots before the model starts.
The formalizer then builds a blueprint, a project-local source map linking source spans to planned \lean declarations, dependencies, proof notes, and recorded scope changes.
Before proof search, a statement/source gate checks that the generated declaration types still match the original source claims.
During proving, a programmatic workflow manager owns the theorem queue, failed-attempt memory, retry budgets, verification records, logs, and checkpoints.
The model proposes edits and helper lemmas, while the manager assigns one target at a time and advances only after external \lean verification accepts the edit.

This separation has different benefits across the two model settings in our experiments.
For \kimi~\citep{moonshot2026kimi}, workflow control is decisive in our two document-level runs: the full workflow finishes both sources, while no-queue variants exhaust the call budget.
For \gpt, all document-level variants finish, so the observed benefit is not a binary success difference; the same structure instead improves token efficiency and preserves an explicit audit trail for the proof work.

\method also introduces LeanProbe\footnote{\url{https://github.com/epfl-lara/LeanProbe}}, a cached same-file verifier built on LeanInteract~\citep{leaninteract2025}, for repeated proof-repair checks.
LeanProbe gives the proving agent low-latency diagnostics and proof-state feedback while leaving final acceptance to standard \lean/Lake checks.
In the sequential same-file benchmark summarized in \cref{tab:leanprobe-summary}, cached checking is roughly 9--14$\times$ faster than rerunning growing-prefix Lake checks.

Our main evaluation ablates \method on real document-level formalizations, where the system must recover statements from source documents and then complete the resulting proof work.
We compare queue management with free-running proof loops, full \lean tool access with terminal-only interaction, and source-blueprint checkpoints with direct source-to-code generation.
Proof-only benchmarks remain useful calibration, but they provide the formal statement in advance and only test whether the system can replace the proof placeholder.
We therefore report RLM25-PFR agent runs and the ICML 2026 AI for Math TCS challenge~\citep{ai4math2026challenge} as complementary measurements.

The two document-level case studies are Frisch and Vaserstein's \emph{Parametrization of Pythagorean Triples by a Single Triple of Polynomials}~\citep{frisch2007pythagorean} and Lyons and Zumbrun's \emph{A Calculus Proof of the Cramer--Wold Theorem}~\citep{lyons2017cramerwold}.
The Pythagorean source combines integer-valued polynomial definitions, an obstruction over $\mathbb{Z}[x_1,\ldots,x_n]$, explicit rational-polynomial witnesses, and custom plain-\TeX{} theorem macros.
The Cramer--Wold source combines probability measures on Euclidean space, closed half-space values, Crofton reconstruction of average-distance functions, odd-dimensional Laplacian inversion, and an even-to-odd embedding step.
These sources are short enough for controlled ablations, but they are still full mathematical papers with nontrivial definitions, representation choices, and proof dependencies.
Both require source-aware definitions and reusable \lean infrastructure that the prose leaves implicit.

This paper makes three \textbf{main contributions}:
\begin{itemize}
    \item A document-to-project workflow for \lean formalization. \method starts from a paper or \TeX{} project, records which source text supports each generated declaration, checks statements before proof search, and moves to the next theorem only after \lean accepts the current proof edit.
    \item A fast local verifier for proof repair. LeanProbe caches the \lean environment before the current declaration and returns proof states and diagnostics through a CLI and MCP server; final acceptance remains a standard \lean/Lake build.
    \item An evaluation on full mathematical sources and proof-only benchmarks. We ablate queue control, tool access, and source checkpoints on two paper formalizations with \kimi and \gpt, and report RLM25-PFR and ICML 2026 AI for Math TCS challenge runs as calibration.
\end{itemize}
We release on GitHub the \href{https://github.com/epfl-lara/LeanFlow}{\method implementation},
\href{https://github.com/epfl-lara/LeanProbe}{LeanProbe verifier}, and
\href{https://github.com/epfl-lara/AutoformalizedProjects}{generated case-study projects}.

\section{Related Work}

\paragraph{Formal theorem proving with language models.}
Lean 4~\citep{demoura2021lean4} and \mathlib~\citep{mathlib2020} provide the verification substrate for many recent neural theorem-proving systems.
Benchmarks such as miniF2F~\citep{zheng2022minif2f}, ProofNet~\citep{azerbayev2023proofnet}, PutnamBench~\citep{tsoukalas2024putnambench}, FormalMATH~\citep{yu2025formalmath}, SorryDB~\citep{letson2026sorrydb}, and VeriSoftBench~\citep{xin2026verisoftbench} measure different proof-only or repository-scale capabilities.
They provide useful calibration, but most start from an existing formal statement or repository context.
Our work instead studies the runtime needed when the source begins as mathematical prose, and the output must be a buildable project.

\paragraph{Autoformalization across scales.}
Early LLM autoformalization work studied natural-language-to-formal translation at theorem scale~\citep{wu2022autoformalization}; later work improved process supervision and evaluation for \lean statements~\citep{lu2024process,poiroux2024reliable,poiroux2025rlmeval}.
M2F~\citep{wang2026m2f} and Automatic Textbook Formalization~\citep{gloeckle2026automatictextbookformalization} formalize long mathematical sources into \lean projects, audit generated statements, and repair proofs under a pinned environment.
Math, Inc.'s Gauss system is an autoformalization agent for large \lean projects~\citep{mathinc2025gauss}. In the sphere-packing project~\citep{hariharan2026milestoneformalizationspherepacking}, it helps complete a sorry-free formalization of the dimension-8 result from an existing blueprint and \lean development.
AlphaProof Nexus~\citep{tsoukalas2026advancingmathematicsresearchaidriven} uses an agentic evolutionary framework for formal proof search over a large set of formalized conjectures.
Direct comparison with M2F is difficult from the currently available public artifacts: reproducing our document-level runs under M2F would require the same source documents, model, \mathlib version, budget, prompts, workflow logs, tool surface, and final acceptance criteria.
Our study is smaller in document volume but different in emphasis: we isolate queue control, tool routing, statement handoff, and model-side budget under the two model settings in our experiments.

\paragraph{Lean tooling for agents.}
Retrieval and environment access are central to practical \lean proving.
LeanDojo~\citep{yang2023leandojo} studies retrieval-augmented theorem proving, and LeanExplore~\citep{asher2025leanexplore} provides search over \lean declarations.
LeanInteract~\citep{leaninteract2025} exposes a Python interface to \lean execution.
\method builds on this tool ecosystem but introduces LeanProbe as its own cached same-file verifier for agent loops; final acceptance still comes from standard \lean/Lake verification.

\section{Problem Setting}

Let $D$ be a mathematical source artifact: a \LaTeX{} file, a PDF, or a directory containing a \TeX{} project.
Let $E$ be a pinned \lean environment: a Lean toolchain, a fixed mathematical-library revision such as a \mathlib commit, and the Lake project configuration used for checking.
The mathematical result should not depend on this particular snapshot, but the evaluation of builds, diagnostics, imports, and available lemmas is always relative to $E$.
The goal is to produce a \lean project $P$ together with declaration-level provenance.
For each generated declaration $d \in \mathrm{Decl}(P)$, the workflow should record
\[
  \pi(d) \subseteq \mathrm{Span}(D),
\]
where $\pi(d)$ is the finite set of source spans, labels, sections, pages, equations, or bibliography references  justifying $d$.

We separate three checks.
Statement type-checking asks whether the generated \lean declaration is well typed under $E$ when theorem-like proof bodies are replaced by explicit \texttt{sorry} placeholders.
Statement faithfulness asks whether that well-typed declaration states the source claim, including statement type, quantifiers, domains, hypotheses, and conclusion.
Proof completion asks whether a faithful declaration has a complete proof with no unacceptable placeholders or custom axioms.
The final project gate is stricter than a theorem-local check: the requested file or full Lake project must build under $E$, and a hygiene scan must report no unapproved \texttt{sorry}, \texttt{admit}, \texttt{unsafe}, or hidden axiom.

\begin{figure*}[t]
\centering
\resizebox{0.98\textwidth}{!}{%
\begin{tikzpicture}[font=\sffamily\scriptsize, node distance=5mm]
  \node[epfStage, text width=24mm] (source) {Source artifact\\\texttt{.tex}, PDF, \TeX{} tree};
  \node[epfStage, text width=24mm, right=of source] (preflight) {Preflight\\manifest\\blocks, PDFs, refs};
  \node[epfStage, text width=24mm, right=of preflight] (blueprint) {Blueprint\\source map\\deps, names, notes};
  \node[epfStage, text width=24mm, right=of blueprint] (draft) {Lean draft\\statements with\\\texttt{sorry} proofs};
  \node[epfGate, text width=24mm, right=of draft] (review) {Statement/source\\gate\\approve or redraft};

  \node[epfStage, text width=28mm, below=18mm of draft] (prove) {Prover queue\\one declaration\\at a time};
  \node[epfTool, text width=28mm, right=of prove] (fast) {LeanProbe\\cached checks\\warm env};
  \node[epfGate, text width=28mm, right=of fast] (build) {Final gates\\Lake build\\sorry/axioms};
  \node[inner sep=0pt, minimum width=40mm, minimum height=15mm, below=9mm of preflight] (state) {};
  \path[
    draw=epfGreen!65,
    fill=epfGreen!5,
    line width=0.55pt
  ]
    ($(state.center)+(-20mm,4.5mm)$)
    arc[start angle=180,end angle=0,x radius=20mm,y radius=3mm]
    -- ($(state.center)+(20mm,-4.5mm)$)
    arc[start angle=0,end angle=-180,x radius=20mm,y radius=3mm]
    -- cycle;
  \draw[draw=epfGreen!65, line width=0.55pt]
    ($(state.center)+(0,4.5mm)$) ellipse[x radius=20mm,y radius=3mm];
  \node[font=\sffamily\scriptsize\bfseries, align=center] at ($(state.center)+(0,-1.9mm)$)
    {Workflow state record};

  \foreach \a/\b in {source/preflight,preflight/blueprint,blueprint/draft,draft/review,prove/fast,fast/build}
    \draw[epfArrow] (\a) -- (\b);
  \coordinate (handoff) at ($(review.south)+(0,-10mm)$);
  \coordinate (handoffProve) at (prove.north |- handoff);
  \draw[epfArrow] (review.south) -- (handoff) -- (handoffProve) -- (prove.north);
  \node[
    font=\sffamily\tiny,
    text=epfOrange,
    fill=white,
    inner sep=0.45mm
  ] at ($(handoff)!0.54!(handoffProve)+(0,2.2mm)$) {Formal statements};
  \coordinate (redraftCeil) at ($(draft.north)!0.5!(review.north)+(0,5.5mm)$);
  \coordinate (redraftRight) at (review.north |- redraftCeil);
  \coordinate (redraftLeft) at (draft.north |- redraftCeil);
  \draw[-{Latex[length=1.7mm]}, draw=epfOrange!80, line width=0.55pt]
    (review.north) -- (redraftRight)
    -- node[midway, above, font=\sffamily\tiny, text=epfOrange, fill=white, inner sep=0.35mm]
      {redraft}
    (redraftLeft) -- (draft.north);
  \coordinate (formalState) at ($(preflight.south)!0.5!(blueprint.south)$);
  \draw[epfStateLink] ($(state.north)+(1.4mm,0)$) -- ($(formalState)+(1.0mm,0)$);
  \draw[epfStateLink] ($(formalState)+(-1.0mm,0)$) -- ($(state.north)+(-1.4mm,0)$);
  \draw[epfStateLink] ($(state.east)+(0,1.2mm)$) -- ($(prove.west)+(0,1.2mm)$);
  \draw[epfStateLink] ($(prove.west)+(0,-1.2mm)$) -- ($(state.east)+(0,-1.2mm)$);

  \node[
    epfGroup=epfBlue,
    fit=(source)(preflight)(blueprint)(draft)(review)(redraftCeil),
    label={[font=\sffamily\scriptsize\bfseries,text=epfBlue]above:Formalizer}
  ] {};
  \node[
    epfGroup=epfTeal,
    fit=(prove)(fast)(build),
    label={[font=\sffamily\scriptsize\bfseries,text=epfTeal]below:Prover}
  ] {};
\end{tikzpicture}
}
\caption{\method document-to-project pipeline. Dashed regions separate the formalizer and prover phases; the green silo is the project-local workflow state record shared by both. The formalizer emits a well-typed statement skeleton and a durable blueprint, and formal theorem statements pass through a statement/source gate before the prover queue attempts proof closure under verifier and hygiene gates. The workflow state records activity, proof checkpoints, failed attempts, route decisions, prover plans, and outcomes, so redrafts and build failures persist outside the transient conversation state.}
\label{fig:pipeline}
\end{figure*}
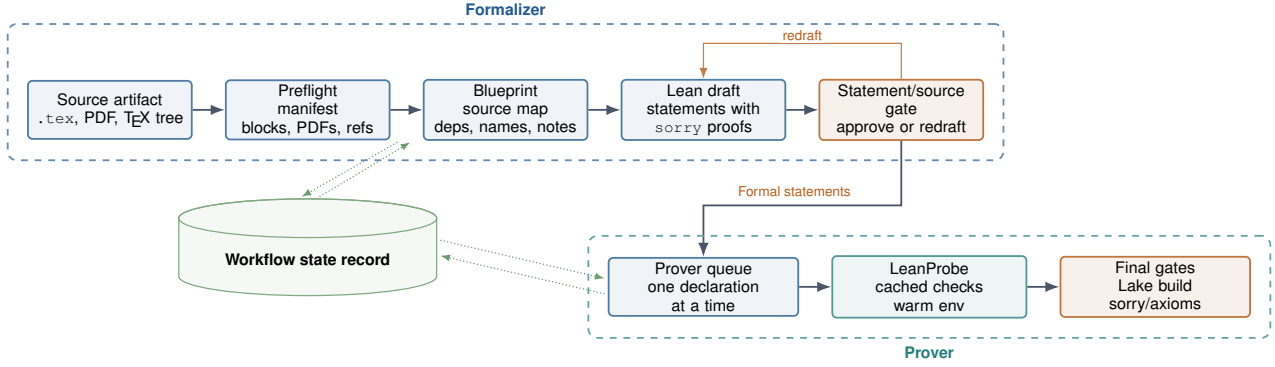

\section{Method}

\subsection{Workflow Overview}

\cref{fig:pipeline} and \cref{alg:method} summarize the document-to-project contract.
\method uses two primary workflows: \texttt{formalize}, which turns a source document into source-backed \lean declarations and proof plans, and \texttt{prove}, which repairs or completes \lean proofs until the requested file or project compiles successfully.
The workflows are deliberately separated.
Formalization may introduce definitions, names, imports, and theorem statements, but it stops at a proof skeleton with source-backed \texttt{sorry} bodies.
Proof repair then treats those declarations as fixed targets: it may edit proof bodies and add local helper lemmas, but it may not modify theorem statements, add axioms, or leave the assigned proof partially solved.

\begin{algorithm}[t]
\caption{\method agent workflow contract}
\label{alg:method}
\begin{algorithmic}[1]
\STATE Resolve source document $D$ and pinned \lean environment $E$.
\STATE Extract a preflight manifest: source blocks, labels, references, PDFs, figures, and support files.
\STATE Create a ``blueprint'' mapping source spans to planned declarations, dependencies, scope changes, and proof notes.
  \STATE Draft \lean declarations with placeholders only for theorem-like proofs; verify that the statement layer type-checks.
\STATE Run a statement/source gate; if faithfulness fails, redraft before proof search.
\WHILE{the requested file or project still contains assigned proof obligations}
  \STATE Assign one theorem-like declaration from the queue.
  \STATE Use search, proof context, and LeanProbe-backed checks to repair the assigned proof.
  \STATE Persist failures, diagnostics, route decisions, and proof checkpoints after each attempt.
  \STATE Accept the assignment only after verification, no relevant \texttt{sorry}, and an acceptable axiom profile.
\ENDWHILE
\end{algorithmic}
\end{algorithm}

\subsection{Source-Backed Skeleton Construction}

\method begins by resolving the source deterministically using document processing tools.
For a \TeX{} project directory, deterministic preflight uses regular expression search to identify and select a main entry point, follow local includes, record bibliography files, figures, PDFs, and support files, and fail on ambiguous roots before the agent starts.
The same pass extracts standard theorem-like environments, custom theorem declarations, plain-\TeX{} theorem/proof blocks, labels, references, citations, section structure, and nearby proof environments when available.
For PDFs, it records extracted text, metadata, and image inventories through local Poppler tools such as \texttt{pdftotext}, \texttt{pdfinfo}, and \texttt{pdfimages}; the model reads this bounded source view through \texttt{read\_pdf}.

The \emph{formalize} workflow next creates a project-local blueprint before the proof skeleton is considered ready.
The blueprint is the durable source map for the run: it lists source statements, planned \lean names, dependencies, source locators, statement-fidelity checks, proof notes, and known scope changes.
For each source theorem or lemma, it spells out the parts of the formal type that are most prone to drift: statement type, quantifier order, parameter domains, codomains, hypotheses, side conditions, and follow-on equivalences.
The generated \lean module must type-check with theorem bodies left as \texttt{by sorry}; proof filling is deferred until the statement/source gate has approved the skeleton.

\subsection{Statement/Source Gate}

The statement/source gate checks the generated skeleton before proof search begins.
Compilation only shows that a declaration is well typed; it does not show that the declaration is the theorem stated in the paper.
\method therefore launches a fresh LLM reviewer context---a separate model context with no human in the loop---over the blueprint, generated \lean declarations, and original source spans.
The reviewer is a single invocation of a thinking model; it can approve, reject, or request redrafting, but it does not fill proof bodies.
If any check in \cref{tab:review-checks} fails, the formalizer returns to the skeleton before the \emph{prove} workflow starts; the prover should not receive a weakened, strengthened, or incomparable target.
We call the version that passes this gate the \emph{reviewed skeleton}.

\begin{table}[t]
\centering
\caption{Statement/source review checks applied before proof repair.}
\label{tab:review-checks}
\footnotesize
\setlength{\tabcolsep}{3pt}
\begin{tabular}{p{0.24\columnwidth}p{0.69\columnwidth}}
\toprule
Check & Rejection signal \\
\midrule
Statement type & The type of the generated \lean declaration does not fully match the source statement, so a proof may certify a different claim \\
Quantifiers & Order, dependency, or implicit parameters changed \\
Variable types & Variables range over the wrong type \\
Hypotheses & Added assumption, missing side condition, changed nonemptiness \\
Conclusion & Weakened equality, image statement, equivalence, or existence \\
Encoding bridge & A local encoding is used without the definitions or equivalence/evaluation lemmas needed to recover the source statement \\
Proof notes & Source proof idea missing for a nontrivial theorem \\
Hygiene & Hidden \texttt{axiom}, \texttt{unsafe}, \texttt{admit}, or unreported \texttt{sorry} \\
\bottomrule
\end{tabular}
\end{table}

\subsection{Project-wide Prover Queue}
The project-wide prover queue chooses which file the prover should work on next.
When a requested project still contains \texttt{sorry}, the manager scans the files, ranks candidate files using declaration dependencies, local hints, examples, and size estimates, records the route decision, and hands one file to the file-scoped declaration queue.
This outer queue is needed because document formalizations produce many dependent proof obligations.
Without route control, the agent can spend calls on the wrong file, revisit failed paths, or move to later work before earlier dependencies are stable.

\subsection{File-Scoped Declaration Queue}

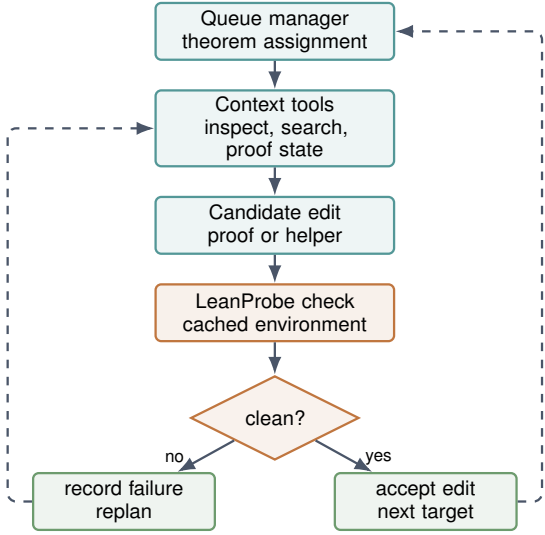
\begin{figure}[t]
\centering
\resizebox{0.86\columnwidth}{!}{%
\begin{tikzpicture}[
  font=\sffamily\scriptsize,
  node distance=3.5mm,
  feedback/.style={epfArrow, dashed, rounded corners=3pt}
]
  \node[epfTool, text width=28mm, minimum height=7mm, inner sep=2pt] (queue) {Queue manager\\theorem assignment};
  \node[epfTool, below=of queue, text width=28mm, minimum height=7mm, inner sep=2pt] (context) {Context tools\\inspect, search, proof state};
  \node[epfTool, below=of context, text width=28mm, minimum height=7mm, inner sep=2pt] (patch) {Candidate edit\\proof or helper};
  \node[epfGate, below=of patch, text width=28mm, minimum height=7mm, inner sep=2pt] (probe) {LeanProbe check\\cached environment};
  \node[epfDecision, below=4mm of probe, text width=15mm] (ok) {clean?};
  \node[epfStore, below left=4mm and 2mm of ok, text width=21mm, minimum height=7mm, inner sep=2pt] (fail) {record failure\\replan};
  \node[epfStore, below right=4mm and 2mm of ok, text width=21mm, minimum height=7mm, inner sep=2pt] (pass) {accept edit\\next target};

  \draw[epfArrow] (queue) -- (context);
  \draw[epfArrow] (context) -- (patch);
  \draw[epfArrow] (patch) -- (probe);
  \draw[epfArrow] (probe) -- (ok);
  \draw[epfArrow] (ok) -- node[pos=.55, left, xshift=-1mm, font=\sffamily\tiny] {no} (fail);
  \draw[epfArrow] (ok) -- node[pos=.55, right, xshift=1mm, font=\sffamily\tiny] {yes} (pass);
  \draw[feedback] (fail.west) -- ++(-3mm,0) |- (context.west);
  \draw[feedback] (pass.east) -- ++(3mm,0) |- (queue.east);
\end{tikzpicture}}
\caption{File-scoped proving loop. Failed checks update theorem-local memory; accepted edits advance to file/Lake verification sweeps.}
\label{fig:prover-loop}
\end{figure}

\cref{fig:prover-loop} shows the declaration queue and repair loop inside an assigned file.
Once the project queue selects a file, the manager builds a file-scoped declaration queue from pending \texttt{sorry} declarations and diagnostics.
The front queue item becomes the active declaration; the prover loop repairs only that declaration before the manager advances to the next queue item.
For the active queue item, the manager stores the target, theorem-local failed attempts and diagnostics, retry counters, and the latest verification record; it also tracks the remaining declarations in the file and each accepted or failed outcome.
It builds the bounded context shown to the model, runs or routes verification, and lets the model advance only after the assigned declaration is clean.
The model sees the current target, a small view of the remaining file work, relevant source/proof context, and theorem-local failed attempts.

The file-scoped declaration queue keeps proof search focused by exposing only one active assignment at a time.
It prevents common long-run failures: losing the active theorem after compaction, retrying a previously failed proof shape, changing a future statement, or advancing because a text search no longer finds \texttt{sorry} while \lean still reports open goals.
After each edit, the proving LLM agent refreshes the structured \lean state and asks the manager whether to continue the same declaration, accept the edit, restore a baseline \texttt{sorry}, or assign the next declaration.

\begin{figure}[bth]
\centering
\begin{leanexamplebox}[title={LeanProbe feedback on a failed candidate}]
\textcolor{leanKeyword}{theorem} add\_comm\_candidate (x y : \textcolor{leanType}{Nat}) :\\
\hspace*{1.0em}x + y = y + x := \textcolor{leanKeyword}{by}
\begin{leanfeedbackbox}
\textcolor{leanComment}{/- \textless{}feedback\textgreater{}}\\
\textcolor{leanComment}{-- proof state:}\\
x y : \textcolor{leanType}{Nat}\\
\textcolor{leanGoal}{$\vdash$} x + y = y + x\\
\textcolor{leanComment}{\textless{}/feedback\textgreater{} -/}
\end{leanfeedbackbox}
\hspace*{1.0em}\textcolor{leanError}{\underline{rfl}}
\begin{leanerrorbox}
\textcolor{leanComment}{/- \textless{}feedback\textgreater{}}\\
\textcolor{leanComment}{--} \textcolor{leanError}{error: tactic 'rfl' failed}\\
\textcolor{leanComment}{\textless{}/feedback\textgreater{} -/}
\end{leanerrorbox}
\end{leanexamplebox}
\caption{Example LeanProbe feedback embedded into a failed \lean candidate. The proof state is injected before the failing tactic (green marker) and the diagnostic after it (red marker); the failing tactic is underlined in red.}
\label{fig:leanprobe-feedback}
\end{figure}

\subsection{LeanProbe and Tool Surface}
\label{sec:leanprobe-tools}

LeanProbe is a declaration-level verifier for repeated proof attempts in one \lean file.
It is a standalone CLI, Python library, and MCP server built on LeanInteract~\citep{leaninteract2025}, and is openly available at \href{https://github.com/epfl-lara/LeanProbe}{\texttt{epfl-lara/LeanProbe}}.
At assignment time, LeanProbe runs a \emph{prepare} step for the active declaration: it checks imports, the file header, and all prior declarations, then stores the resulting environment immediately before the target.
Each \emph{check} step replaces the target declaration with one candidate proof or helper block and checks it against that cached environment.
The cache advances only after a complete declaration replacement is accepted, so failed candidates give diagnostics without changing the environment seen by later obligations.

LeanProbe returns proof-agent feedback: diagnostics, warnings, \texttt{sorry} detection, tactic metadata, goal states, and inline \lean comments at the failing location.
A failed candidate can be returned as model-readable \lean with compact comments containing the local proof state and the diagnostic beside the failing tactic, as illustrated in \cref{fig:leanprobe-feedback}.

\begin{table}[bth]
\centering
\caption{Default tool surface exposed to the agent currently comprises 27 tools. The following table groups them by purpose. The manager decides which tool class is relevant for the current proof obligation or failure mode.}
\label{tab:toolset}
\scriptsize
\setlength{\tabcolsep}{3pt}
\begin{tabular}{p{21mm}p{54mm}}
\toprule
Tool class & Agent use \\
\midrule
Source readers & Inspect \TeX/PDF inputs, extracted text, image inventory, and source spans \\
File editing & Read, search, patch, and write project files under manager checks \\
Lean verification & Run canonical \lean/Lake checks and LeanProbe cached target checks \\
Search/context & Retrieve local/\mathlib facts, nearby declarations, and theorem context \\
Proof exploration & Try bounded proof candidates, request reasoning help, and decompose helpers \\
Code hygiene & Scan for remaining \texttt{sorry}, custom axioms, unsafe constructs, and axiom profiles \\
Workflow memory & Reattach skills, blueprints, session notes, and theorem-local failed attempts \\
Terminal fallback & Run explicit shell commands when the bounded tools are insufficient \\
\bottomrule
\end{tabular}
\end{table}

This information is used for fast local repair inside the queue loop, not for final acceptance.
The final gates remain module or project Lake builds and verification that the project is free from unapproved \texttt{axiom} declarations and remaining \texttt{sorry} placeholders.
\cref{tab:toolset} summarizes the tool classes exposed to the model; ablations in \cref{sec:experiments} remove this surface to isolate its effect.

\section{Evaluation}
\label{sec:experiments}

The experiments test whether the \method workflow components in \cref{fig:pipeline} improve real formalization and proving runs.
Each completed run reports task outcome, API calls, input tokens, and output tokens from the workflow logs; the released \method artifacts include the run configurations, prompts, and logs used for these tables.
All model-backed runs use fixed decoding settings across conditions: temperature $0.3$, top-$p=0.95$, and a maximum output budget of 65{,}536 tokens per model turn.
We count only model API calls.
For \gpt, one Codex CLI model turn is one call; for \kimi, one request to the EPFL-hosted inference endpoint is one call.
Terminal commands, file reads and edits, \lean/Lake invocations, LeanProbe checks, and other structured tool invocations are not counted as API calls.
For cost accounting, the \gpt runs used the Codex CLI under a US\$200 GPT Pro subscription; one month of this subscription covered all \gpt experiments reported here.
The \kimi runs used a self-hosted EPFL deployment on $8\times$H200 GPUs, and token-cost estimates use CHF 0.4802 per million input tokens and CHF 1.4406 per million output tokens.
Because these two models use different accounting regimes, the tables report calls and tokens rather than converting all rows to a single monetary scale.

\subsection{Document-Level Case Studies}

The two document-level case studies are selected for mathematical content and area diversity.
Frisch and Vaserstein's Pythagorean-polynomial paper~\citep{frisch2007pythagorean} requires formalizing the classical $T(a,b,c)$ parametrization, parity and divisibility side conditions, the obstruction to a single integer-coefficient polynomial triple, and the explicit four-variable triple of integer-valued rational polynomials.
Lyons and Zumbrun's Cramer--Wold paper~\citep{lyons2017cramerwold} requires formalizing closed half-spaces as measurable sets, equality of measures from matching half-space values, the Crofton step from half-space values to average-distance functions, the odd-dimensional Laplacian recovery argument, and the even-to-odd embedding $\mathbb{R}^{2m}\hookrightarrow\mathbb{R}^{2m+1}$.
\cref{app:case-studies} gives the detailed source-to-\lean decomposition for both case studies.
To the best of our knowledge, neither source theorem had a prior proof-assistant formalization.
The completed Pythagorean project comprises 83 \lean declarations (34 of them \texttt{theorem}/\texttt{lemma} statements) and the Cramer--Wold project 114 declarations (101 \texttt{theorem}/\texttt{lemma} statements); both build sorry-free and free of unapproved axioms.

For the workflow ablations reported in \cref{tab:longform-ablation}, each source first runs through the pre-proof stages in \cref{fig:pipeline,alg:method}: source resolution, blueprint construction, statement generation with proof placeholders, and statement/source review.
The result is a reviewed skeleton, i.e., a \lean statement layer that type-checks with theorem bodies left as placeholders and has passed the source-faithfulness checks in \cref{tab:review-checks}.
The Pythagorean reviewed skeleton is produced with \kimi; the Cramer--Wold reviewed skeleton is produced with \gpt because the initial representation setup for probability measures, half-spaces, and the analytic inversion target requires more representation choices than the Pythagorean source.
These pre-proof stages are held fixed across the proof-repair ablations, and the 2000-call budget in \cref{tab:longform-ablation} begins after the reviewed skeleton has been accepted.
Statement translation, skeleton construction, the statement/source gate, and proof repair all run autonomously, with no human edits to the reviewed skeletons or to the proof runs in \cref{tab:longform-ablation}; expert involvement is limited to verifying faithfulness after the fact (\cref{sec:threats}), because no automated faithfulness guarantee is currently available.

\subsection{Workflow Ablations on Real Sources}

The primary evaluation is an ablation study: it compares the full \method against variants that remove the queue system and/or the specialized \lean tools.
For each model, we run the full $2\times2$ design: queue versus no queue, and full tool surface versus terminal-only interaction. \cref{tab:longform-ablation} is organized by source so that completion behavior and model-side budget can be compared within the same mathematical project.
Here and in later tables, \emph{queue} means the two-layer queue system: the project-wide file queue plus the file-scoped declaration queue.
All four conditions for a given source start from the same reviewed skeleton and use the same proof goal, prompt template, model settings, and call budget; only the queue system and tool surface are changed.
All document-level proving runs have a maximum budget of 2000 prover-agent API calls, where one call is one model turn in the proof-repair loop.
In the terminal-only condition, the agent may still read and edit files, search with shell tools such as \texttt{rg}, and run commands such as \texttt{lake build}, but it has no access to LeanProbe, MCP-backed \lean tools, structured declaration search, or proof-state APIs.
In the no-queue condition, an outer runner still keeps the prover running until the project succeeds or the call budget expires, but both queue layers are disabled: the manager does not route files through a project queue, does not restrict the model to one top-of-queue declaration, and does not use theorem-local failed-attempt history to gate the next target.
Source preflight, the statement/source gate, final \lean/Lake verification, and hygiene checks are held fixed across conditions.

\begin{table*}[t]
\centering
\caption{Long-form workflow ablation on the two document-level case studies. Workflow abbreviations encode queue-system and tool configuration: Full = both queue layers with full tools; NoQ+Tools = no queue system with full tools; Q+CLI = both queue layers with terminal-only interaction; NoQ+CLI = no queue system with terminal-only interaction. Outcome is \emph{success} when the requested project builds under $E$ from the reviewed skeleton with no remaining \texttt{sorry} and no unapproved axioms within the 2000-call budget, and \emph{failure} otherwise. Calls are prover-agent API calls with a 2000-call cap.}
\label{tab:longform-ablation}
\footnotesize
\setlength{\tabcolsep}{3pt}
\begin{tabular*}{\textwidth}{@{\extracolsep{\fill}}llllrrr@{}}
\toprule
Source & Model & Workflow & Outcome & Calls & In tok. & Out tok. \\
\midrule
Pythagorean & \kimi & Full & \textbf{success} & 1043 & 46.9M & 514.1K \\
Pythagorean & \kimi & NoQ+Tools & failure & 2000 & 160.2M & 682.0K \\
Pythagorean & \kimi & Q+CLI & \textbf{success} & 1561 & 91.3M & 537.8K \\
Pythagorean & \kimi & NoQ+CLI & failure & 2000 & 157.5M &  615.5K \\
Cramer--Wold & \kimi & Full & \textbf{success} & 1278 & 66.0M & 684.1K \\
Cramer--Wold & \kimi & NoQ+Tools & failure & 2000 & 127.8M & 583.2K \\
Cramer--Wold & \kimi & Q+CLI & failure & 2000 & 113.1M & 579.0K \\
Cramer--Wold & \kimi & NoQ+CLI & failure & 2000 & 167.0M  & 496.4K \\
\midrule
Pythagorean & \gpt & Full & \textbf{success} & 258 & 14.3M & 190.5K \\
Pythagorean & \gpt & NoQ+Tools & \textbf{success} & 568 & 45.0M & 240.3K \\
Pythagorean & \gpt & Q+CLI & \textbf{success} & 277 & 14.3M & 177.0K \\
Pythagorean & \gpt & NoQ+CLI & \textbf{success} & 984 & 89.9M & 340.7K \\
Cramer--Wold & \gpt & Full & \textbf{success} & 495 & 25.9M  & 305.5K\\
Cramer--Wold & \gpt & NoQ+Tools & \textbf{success} & 448 & 32.8M & 259.7K \\
Cramer--Wold & \gpt & Q+CLI & \textbf{success} & 959 & 59.5M & 518.0K \\
Cramer--Wold & \gpt & NoQ+CLI & \textbf{success} & 476 & 49.7M & 281.6K \\
\bottomrule
\end{tabular*}
\end{table*}

\cref{tab:longform-ablation} shows that the combined queue system is decisive for \kimi in these two document-level runs: both no-queue variants exhaust the 2000-call budget on both case studies, while full \method succeeds in 1043 calls on Pythagorean triples and 1278 calls on Cramer--Wold.
The effect of the tool surface is most visible on the analytic case study, where \kimi with a queue but terminal-only interaction still fails on Cramer--Wold.
For \gpt, all document-level variants succeed, so these rows do not establish \method as necessary for completion under this model. Instead, they support an efficiency and auditability claim: full \method has the lowest or tied-lowest input-token cost on both sources, although the Cramer--Wold no-queue/full-tools run uses slightly fewer calls at a higher input-token cost.

\subsection{RLM25-PFR Autoformalization}

We evaluate on RLM25-PFR, the PFR slice of RLM25~\citep{poiroux2025rlmeval,poiroux2024reliable}, rather than on the full RLM25 benchmark.
This slice contains 145 theorem-statement examples from the Polynomial Freiman--Ruzsa project.
We report it separately from the document-to-project case studies because each example is an independent theorem task rather than a complete source document.

Each RLM25-PFR example is run as a two-stage \lean workflow with one shared 200-call model budget per example.
First, the agent receives the informal statement, project context, and any available informal proof or blueprint material, and synthesizes a \lean declaration that must type-check, typically with a placeholder proof.
The generated statement is evaluated against the reference formal statement using the benchmark's BEq and BEq+ statement-equivalence metrics.
Second, the generated statement is treated as a fixed proof target: the prover may fill proof bodies and add local helper lemmas, but it may not weaken statements, introduce custom axioms, or leave \texttt{sorry}/\texttt{admit} placeholders.
The call budget is shared by statement synthesis and proof completion, and the call totals in \cref{tab:autoformalization-experiments} include both stages.
Runs are autonomous after launch, with no human edits to the generated statements or proofs.
If the 200-call budget is exhausted before final verification, the example is counted as a proof failure.
Proof success requires final \lean/Lake verification and the same hygiene scan used in our document-level runs.
\cref{tab:autoformalization-experiments} summarizes the completed \gpt runs. In this single \gpt run, \method reduces calls (3{,}541 vs.\ 4{,}053) and input tokens (203.2M vs.\ 236.6M) while increasing BEq+ from 72.9\% to 75.7\%; because the runs lack variance estimates, we treat the BEq+ change as a small effect rather than a significance claim. The gap between proof success and statement-equivalence metrics reinforces why verified proof completion cannot replace statement/source checks: a proved declaration can still formalize the wrong mathematical claim.

To our knowledge, no published baseline reports provide proof-of-completion results for the RLM25-PFR two-stage setting; the original source \cite{poiroux2025rlmeval} is not concerned with producing proofs of statements.

\begin{table}[t]
\centering
\caption{RLM25-PFR agent-run results on the PFR slice. Proof success is the share of examples whose final \lean code verifies with no remaining placeholders or unapproved axioms under the 200-call per-example budget; BEq and BEq+ are statement-equivalence metrics against reference formal statements. Calls and tokens are totals over the slice and include both statement synthesis and proof completion.}
\label{tab:autoformalization-experiments}
\resizebox{\columnwidth}{!}{%
\scriptsize
\setlength{\tabcolsep}{2.5pt}
\begin{tabular}{llllllll}
\toprule
Model & Workflow & Proof succ. & BEq & BEq+ & Calls & In tok. & Out tok. \\
\midrule
\gpt & \method & 81.2\% & 70.8\% & 75.7\% & 3541 & 203.2M & 2.1M \\
\gpt & terminal-only agent & 80.6\% & 68.1\% & 72.9\% & 4053 & 236.6M & 2.3M \\
\bottomrule
\end{tabular}}
\end{table}

\subsection{Challenge Tasks}

Project-level proving tasks provide the formal statements and isolate proof construction.
The ICML 2026 AI for Math Track 2 TCS proving challenge tests project-level algorithmic proof tasks in graph theory, combinatorics, computational complexity, and algorithm correctness using CSLib and a comparator-backed pass-rate metric~\citep{ai4math2026challenge}.
The five challenge projects cover binary heap operations, a heap-backed Dijkstra proof, Kruskal minimum spanning trees, segment-tree construction and range queries, and treap invariant/runtime obligations.
Each run uses the same 2000-call proof budget as the document-level proof-repair runs.
Success means that the final project passes the official SafeVerify/comparator check and also passes our local final \lean/Lake verification and hygiene scan with no remaining \texttt{sorry}/\texttt{admit} placeholders or unapproved axioms.
We run all five projects with \gpt under three conditions: full \method, no queue system with LeanProbe still available, and no queue system with only terminal \lean/Lake interaction.
For \kimi, we additionally run the Kruskal project with full \method and without either the queue system or the specialized tool surface.
All challenge runs are autonomous after launch, with no human edits to the proof files during the run.
The public Track 2 leaderboard lists our competition submission under the \texttt{lmilikic} identifier at rank 2 in the combined Phase 1+2 ranking~\citep{ai4math2026challenge}.

\begin{table*}[!b]
\centering
\caption{AI4Math TCS challenge outcomes by project and workflow condition. Workflow abbreviations encode queue-system and tool configuration: Full = both queue layers with full tools; NoQ+Probe = no queue system with LeanProbe available; NoQ+CLI = no queue system with terminal-only \lean/Lake interaction. Calls are prover-agent API calls with a 2000-call per-project cap.}
\label{tab:proof-experiments}
\footnotesize
\setlength{\tabcolsep}{3pt}
\begin{tabular*}{\textwidth}{@{\extracolsep{\fill}}llllrrr@{}}
\toprule
Project & Model & Workflow & Outcome & Calls & In tok. & Out tok. \\
\midrule
Binary heap & \gpt & Full & \textbf{success} & 54 & 2.86M & 49.8K \\
Binary heap & \gpt & NoQ+Probe & \textbf{success} & 27 & 741.3K & 7.1K \\
Binary heap & \gpt & NoQ+CLI & \textbf{success} & 72 & 5.03M & 71.1K \\
\midrule
Dijkstra & \gpt & Full & \textbf{success} & 61 & 4.04M & 75.1K \\
Dijkstra & \gpt & NoQ+Probe & \textbf{success} & 20 & 770.2K & 26.0K \\
Dijkstra & \gpt & NoQ+CLI & \textbf{success} & 29 & 1.15M & 23.9K \\
\midrule
Segment tree & \gpt & Full & \textbf{success} & 62 & 4.01M & 35.0K \\
Segment tree & \gpt & NoQ+Probe & \textbf{success} & 85 & 7.01M & 42.2K \\
Segment tree & \gpt & NoQ+CLI & \textbf{success} & 73 & 4.67M & 45.4K \\
\midrule
Treap & \gpt & Full & \textbf{success} & 142 & 10.88M & 85.1K \\
Treap & \gpt & NoQ+Probe & \textbf{success} & 144 & 10.80M & 95.6K \\
Treap & \gpt & NoQ+CLI & \textbf{success} & 188 & 14.10M & 116.8K \\
\midrule
Kruskal & \gpt & Full & \textbf{success} & 47 & 1.75M & 16.2K \\
Kruskal & \gpt & NoQ+Probe & \textbf{success} & 37 & 1.56M & 17.7K \\
Kruskal & \gpt & NoQ+CLI & \textbf{success} & 53 & 2.94M & 26.8K \\
\midrule
Kruskal & \kimi & Full & \textbf{success} & 942 & 107.9M & 3.1M \\
Kruskal & \kimi & NoQ+CLI & failure & 2000 & 189.1M & 930.4K \\
\bottomrule
\end{tabular*}
\end{table*}

\cref{tab:proof-experiments} shows that all \gpt variants solve all five TCS projects, so these rows do not isolate a single decisive workflow component.
The results are also consistent with LeanProbe being useful as a feedback surface: workflow logs show that \gpt used LeanProbe in 37\% of recorded agent steps in TCS runs where the tool was available, and NoQ+Probe uses fewer calls and lower input tokens than NoQ+CLI on four of five projects.
This aligns with the tool design in \cref{sec:leanprobe-tools}, where failed candidates can return local compiler diagnostics and proof-state feedback before the next proof attempt.
For \kimi, the tested Kruskal run succeeds with the full workflow, while the no-queue terminal-only variant reaches the 2000-call budget.

\subsection{LeanProbe Utility}

LeanProbe is evaluated separately because it changes verifier latency rather than model capability.
We compare cached LeanProbe checks against terminal \texttt{lake env lean} checks on two workloads that match the proof loop in \cref{fig:prover-loop}.
The repeated-target workload measures many candidate replacements for one active declaration after a single prepare step.
The sequential workload measures a queue-style file run: partial candidates are rejected against the current cached environment, while accepted declarations advance the environment for later targets.

\begin{table*}[t]
\centering
\caption{LeanProbe latency benchmarks. Repeated-target times are average ranges in seconds after one prepare step; sequential rows report total seconds per compact file with five declarations and ten partial/full candidate scenarios.}
\label{tab:leanprobe-summary}
\footnotesize
\setlength{\tabcolsep}{3pt}
\begin{tabular}{@{}llll@{\hspace{7mm}}lllll@{}}
\toprule
Workload & OS & Full-file Lake & Cached check &
Workload & OS & Growing-prefix Lake & Cached total & Speedup \\
\midrule
TCS & macOS & 2.082--2.617s & 0.031--0.049s &
Compact files & macOS & 36.474--67.992s & 3.789--4.775s & 9.63--14.24$\times$ \\
TCS & Linux & 1.495--1.886s & 0.032--0.054s &
Compact files & Linux & 22.593--23.186s & 2.301--2.547s & 9.05--9.82$\times$ \\
\bottomrule
\end{tabular}
\end{table*}

\cref{tab:leanprobe-summary} supports the design choice of checking candidate edits frequently instead of waiting for large repair batches.
The prepare step is paid once before a target, and then each candidate proof replacement is checked in milliseconds to tens of milliseconds.
Sequential checks also match the queue design: the environment advances only after a complete declaration succeeds, so failed candidates are diagnosed without changing the context for later obligations.
The detailed LeanProbe experimental setup and per-file results are reported in \cref{app:leanprobe}.

\subsection{Statement-Skeleton Case Study}

We use a qualitative Pythagorean-polynomial comparison to check whether the source-to-statement stages in \cref{alg:method} prevent statement drift before proof search.
The no-review condition skips the blueprint and statement/source review stages in Steps 3 and 5: after seeing the source, the model is asked to produce \lean translations directly, without a recorded source-to-declaration map or an independent faithfulness check before proving.
The no-review draft and the reviewed skeleton both build, but they formalize different mathematical objects.
The no-review draft would support a less source-faithful function-level formalization of the result: it defines the main witnesses as rational-valued functions on integer inputs.
The reviewed skeleton is the source-faithful version under the checks in \cref{tab:review-checks}: it defines the witnesses as rational multivariate polynomials and separately proves integer-valuedness, matching the paper's theorem.
The detailed declaration-level contrast is reported in Appendix~\ref{app:docs-case-study} (\cref{tab:pyth-skeleton-comparison}).

\section{Threats to Validity}
\label{sec:threats}

The evaluation has several limitations.
Buildability is necessary but not sufficient: a project can build while encoding a different claim.
The statement/source gate runs autonomously, but we still confirm faithfulness by expert inspection, because there is currently no automated way to guarantee that a well-typed declaration matches the source; reducing this manual confirmation is important future work.
The workflow also assumes that the source definitions and theorem statements are the targets to preserve.
After the statement/source gate accepts a skeleton, the prover is not allowed to change definitions or theorem statements; this improves auditability, but it can make a proof obligation impossible if the source contains a typo, an omitted side condition, or an undetected mistranslation.
Such errors must be caught during statement/source review or expert audit rather than silently repaired during proof search.
The document-level evidence covers only two case studies; the Cramer--Wold skeleton was initialized with \gpt rather than \kimi because of its representation choices.
The ablations are single runs, so they do not estimate variance; repeated document-level runs would be more informative but costly, since each can require hundreds to thousands of calls, tens to hundreds of millions of input tokens, and substantial wall-clock time.
Several mechanisms also remain bundled: queue/tool ablations do not isolate blueprinting, statement review, failed-attempt memory, hygiene scanning, and LeanProbe feedback.
The complementary benchmarks do not fully exercise document-to-project formalization, the RLM25-PFR rows cover only the PFR slice rather than the full RLM25 benchmark, the \gpt rows support efficiency claims more than necessity claims, and we do not include a matched document-level M2F comparison.
Calls and token counts are useful proxies for model-side effort, but not full measures of wall-clock, monetary, or human engineering cost.

\section{Conclusion}

\method treats autoformalization as an evolving \lean project with durable workflow state, not as independent theorem prompts.
The central design choice is to keep project control outside the model: the runtime records source-to-\lean decisions, reviews statements before proof search, assigns proofs one at a time, checks candidate edits quickly, and accepts only builds that pass the verifier and contain no leftover placeholders or unapproved axioms.
The document-level evidence is model-dependent: in our two case studies, workflow control is decisive for \kimi completion under budget, while for \gpt it mainly improves token efficiency and audit discipline because every document-level variant completes.

\section*{Impact Statement}

This work aims to make \lean autoformalization more inspectable, reproducible, and verifier-grounded.
Potential positive impacts include reducing the engineering burden of proof repair, improving the auditability of machine-assisted formalization, and making document-level formal verification workflows easier to study.
Potential risks include over-reliance on generated formal artifacts, mistaking buildability for source faithfulness, and unequal access to the compute and model resources needed for long formalization runs.
\method is designed to mitigate these risks by recording source-to-\lean decisions, separating statement review from proof search, preserving workflow logs and checkpoints, and accepting final artifacts only after external \lean verification with no leftover placeholders or unapproved axioms.
The system is intended to assist expert formalizers rather than replace human mathematical judgment.

\section*{Author Contributions}

Lazar Miliki\'c led the system design, implementation, experiments, and paper writing.
Simon Guilloud helped with benchmarking code, research direction, and paper writing. 
Khanh Nguyen contributed with research direction and paper writing.
Viktor Kun\v{c}ak supervised the project, contributed to the research direction, and helped revise the manuscript.

\section*{Acknowledgments}

This research is supported by the AI For Math Fund, administered by Renaissance Philanthropy and funded by XTX Markets. We also gratefully acknowledge the support of the EPFL School of Computer and Communication Sciences.

\bibliographystyle{icml2026}
\bibliography{leanflow_paper}

\clearpage
\onecolumn
\appendix
\section{Document Case Study Details}
\label{app:case-studies}

The \method implementation is available on GitHub at \href{https://github.com/epfl-lara/LeanFlow}{\texttt{epfl-lara/LeanFlow}}, the LeanProbe verifier at \href{https://github.com/epfl-lara/LeanProbe}{\texttt{epfl-lara/LeanProbe}}, and the generated formalization projects at \href{https://github.com/epfl-lara/AutoformalizedProjects}{\texttt{epfl-lara/AutoformalizedProjects}}.
Each released project is self-contained and pins its own \lean toolchain and \mathlib revision (for example, \lean \texttt{v4.30.0-rc2} for the Pythagorean project and \texttt{v4.29.1} for Cramer--Wold), so builds are reproducible without a single global environment.
The two case studies have also been contributed to the community \href{https://github.com/Vilin97/lean-pool}{\texttt{lean-pool}} repository of AI-produced \lean developments (pull requests \href{https://github.com/Vilin97/lean-pool/pull/185}{\#185} and \href{https://github.com/Vilin97/lean-pool/pull/186}{\#186}).

\paragraph{Pythagorean-polynomial source.} \label{app:docs-case-study}
The Frisch--Vaserstein source first defines Pythagorean triples, polynomial parametrization over $\mathbb{Z}[x_1,\ldots,x_n]$, and integer-valued parametrization over $\mathrm{Int}(\mathbb{Z}^n)\subseteq \mathbb{Q}[x_1,\ldots,x_n]$.
The formalization therefore introduces both integer-coefficient and rational-polynomial representations before stating the main claims.
The source-level proof obligations are: the standard two-family parametrization of triples, the obstruction to a single integer-coefficient triple, the explicit four-variable integer-valued witness, the positive-triple variant, and the four-square unrestricted-parameter variant.
The obstruction proof is algebraic: it uses the UFD structure of $\mathbb{Z}[x_1,\ldots,x_n]$, a gcd decomposition, parity facts, and the examples $(3,4,5)$ and $(4,3,5)$ to force a contradiction.
The construction proof uses the source map
\[
  T(a,b,c)=\left(c(a^2-b^2)/2,\;cab,\;c(a^2+b^2)/2\right)
\]
and the parity condition that makes all three coordinates integral.
The reviewed \lean layout separates these concerns into basic definitions, source handoff lemmas, obstruction lemmas, integer-valued construction lemmas, positive-triple lemmas, and explanatory material for cited background.
This layout also records source claims that are likely to be omitted in a proof-only draft: the finite-cover theorem for integer-valued parametrizations, the displayed binomial-polynomial identity, and the non-UFD motivation for $\mathrm{Int}(\mathbb{Z})$.

\begin{table}[H]
\centering
\caption{Declaration-level contrast in the Pythagorean statement skeleton. The no-review draft is generated directly from the source; the reviewed skeleton is the version accepted by the statement/source review conditions described in \cref{tab:review-checks}.}
\label{tab:pyth-skeleton-comparison}
\footnotesize
\setlength{\tabcolsep}{4pt}
\begin{tabular}{@{}p{0.16\textwidth}p{0.39\textwidth}p{0.39\textwidth}@{}}
\toprule
Source requirement & No-review draft & Reviewed skeleton \\
\midrule
Integer-valued polynomial objects &
\texttt{IntegerValued4} is a predicate definition on functions $\mathbb{Z}\to\mathbb{Z}\to\mathbb{Z}\to\mathbb{Z}\to\mathbb{Q}$; integer-valuedness is checked only after evaluation at integer inputs. &
\texttt{RatPoly n} is a type abbreviation for rational multivariate polynomials; \texttt{IsIntValued} is a predicate definition, and \texttt{IntValuedSubring n} is a subring definition. \\
\midrule
Displayed witnesses are polynomials &
\texttt{f\_param}, \texttt{g\_param}, and \texttt{h\_param} are function definitions of four integer arguments, so the displayed source witnesses are encoded only by their evaluations. &
\texttt{f\_param}, \texttt{g\_param}, and \texttt{h\_param} are polynomial definitions of type \texttt{RatPoly 4}; evaluation at integer tuples is handled separately by \texttt{ratPolyEval}. \\
\midrule
Main theorem type &
\begin{tabular}[t]{@{}l@{}}
\texttt{integer\_valued\_}\\
\texttt{parametrization}
\end{tabular}
is a theorem proving integer-valuedness and image equality for the function model. &
\begin{tabular}[t]{@{}l@{}}
\texttt{exists\_int\_valued\_}\\
\texttt{parametrization}
\end{tabular}
is a theorem with conclusion \texttt{IntValuedParametrizes ...}, so it states existence of integer-valued rational polynomials that parametrize the triples. \\
\midrule
Background claims retained &
The direct translation omits theorem or definition declarations for the finite-cover result and the integer-valued-polynomial ring background. &
The reviewed skeleton keeps theorem declarations for the finite-cover result and non-UFD claim, plus polynomial definitions and theorems for the binomial-polynomial identity. \\
\bottomrule
\end{tabular}
\end{table}

\paragraph{Cramer--Wold source.}
The Lyons--Zumbrun source states that Borel probability measures on $\mathbb{R}^n$ are determined by their values on closed half-spaces.
The formalization represents $\mathbb{R}^n$ as \lean Euclidean space, defines closed half-spaces by a normal vector and threshold, and records half-space value functions and the average-distance function
\[
  f_\mu(y)=\int(\|y-x\|-\|x\|)\,d\mu(x).
\]
The first analytic proof block packages the Crofton/Fubini argument showing that half-space values determine $f_\mu$.
The second proof block packages the odd-dimensional inversion step: after reindexing dimensions as $2m+1$, the needed kernel identity is that an $(m+1)$-fold Laplacian of the norm kernel recovers a nonzero multiple of the point mass at the origin.
In \lean-facing terms, this requires a tempered-distribution norm kernel, Fourier/Laplacian algebra for iterated Laplacians, transport between the Euclidean model used by PhysLib~\citep{physlean2025} and the project's odd-dimensional space, and a bridge from PhysLib's real distribution API to Mathlib's complex tempered distributions.
The analytic formalization also produced reusable PhysLib infrastructure for distributional Laplacians of norm-kernel fundamental solutions, contributed upstream in \href{https://github.com/leanprover-community/physlib/pull/1175}{\texttt{leanprover-community/physlib\#1175}}.
The even-dimensional Cramer--Wold theorem is then reduced to the odd-dimensional one by embedding $\mathbb{R}^{2m}$ into $\mathbb{R}^{2m+1}$ and transporting half-space values along that embedding.
The reviewed skeleton keeps the Crofton and inversion blocks as named proof targets rather than folding them into the final theorem; this makes clear which missing analytic facts are mathematical infrastructure, not model-generated assumptions.

\section{LeanProbe Benchmark Results}
\label{app:leanprobe}

This appendix records the detailed LeanProbe benchmark rows behind \cref{tab:leanprobe-summary}.
The snapshot was refreshed on May 13, 2026, with Lean 4.30.0-rc2.
These rows are verifier-latency measurements, not model-performance measurements: they isolate the cost of checking candidate Lean edits under different reuse policies.
The benchmark suite has two parts: grouped repeated-target checks, which stress many alternatives for one active declaration, and sequential same-file checks, which stress queue-style advancement through a file.
All runs use one measured run per target/file, no benchmark warmups, warm Lake caches from prior validation, and a fresh-server baseline.
LeanProbe remains a feedback surface; final file or project acceptance still uses standard \lean/Lake verification.

\paragraph{Grouped repeated-target benchmark.}
The repeated-target workload measures the agent pattern of trying several complete replacements for one active declaration.
For each target, the benchmark writes a temporary full file and times terminal \texttt{lake env lean}; then it starts LeanProbe, prepares the environment immediately before the target, checks the replacement against that cached environment, optionally requests proof-state feedback, and finally repeats the check with a fresh LeanProbe/LeanInteract server to measure the no-cache baseline.
The compact groups are Mathlib-facing examples: \texttt{analysis\_real} covers elementary real-analysis facts, \texttt{algebra\_order} covers ordered algebra and inequalities, \texttt{sets\_functions} covers set/function image and preimage reasoning, and \texttt{number\_theory\_nat} covers elementary natural-number arithmetic.
The TCS groups are longer challenge-style extracts: \texttt{tcs\_binary\_heap} contains heap operations such as \texttt{heapify}, \texttt{extract\_min}, \texttt{insert}, \texttt{merge}, and \texttt{remove}; \texttt{tcs\_treap\_analysis} contains the two probability-sum obligations used in the treap analysis; and \texttt{tcs\_weighted\_graph} contains weighted-graph helper declarations through the \texttt{Sym2order} prefix.
In \cref{tab:appendix-leanprobe-repeated}, \emph{Targets} is the number of active declarations in the group, \emph{Lake full} is full-file terminal verification, \emph{Prepare} is the one-time cached-environment setup, \emph{Cached} is a target replacement check after prepare, \emph{Feedback} includes diagnostic/proof-state metadata, \emph{Fresh} is the same check with a fresh server, and \emph{Fresh/cached} is the fresh-check time divided by the cached-check time.

\paragraph{Sequential same-file benchmark.}
The sequential workload models a queue-managed file run rather than repeated attempts on one target.
Each compact file has five declarations and ten scenarios: for every targetable declaration, the benchmark checks a partial declaration containing \texttt{sorry} and then the complete declaration.
LeanProbe reports the partial scenario and detects the \texttt{sorry}, but it advances the cached environment only after the complete declaration succeeds.
The Lake baselines rerun terminal checks for every scenario, either on a growing prefix that includes accepted prior declarations or on a full temporary file in which only the current declaration is replaced.
In \cref{tab:appendix-leanprobe-sequential}, \emph{Lake prefix} and \emph{Lake full} are these two terminal baselines, \emph{Probe cached} is one warm LeanProbe/LeanInteract server walking the whole file, \emph{Probe fresh} restarts LeanProbe for every scenario, and the three speedup columns divide the corresponding baseline total by the cached LeanProbe total.

\begin{table}[H]
\centering
\caption{LeanProbe repeated-target grouped results. Times are averages in seconds; fresh/cached is the ratio between a fresh LeanProbe server and a cached check.}
\label{tab:appendix-leanprobe-repeated}
\footnotesize
\setlength{\tabcolsep}{3pt}
\begin{tabular}{llrrrrrrr}
\toprule
Platform & Group & Targets & Lake full & Prepare & Cached & Feedback & Fresh & Fresh/cached \\
\midrule
macOS & analysis\_real & 5 & 3.893 & 6.024 & 0.039 & 0.022 & 4.014 & 139.7$\times$ \\
macOS & algebra\_order & 5 & 3.900 & 3.683 & 0.048 & 0.039 & 3.987 & 106.7$\times$ \\
macOS & sets\_functions & 5 & 3.708 & 3.502 & 0.008 & 0.007 & 3.766 & 454.7$\times$ \\
macOS & number\_theory\_nat & 5 & 3.731 & 3.478 & 0.011 & 0.006 & 3.776 & 420.0$\times$ \\
Linux & analysis\_real & 5 & 2.276 & 2.315 & 0.025 & 0.024 & 2.412 & 103.2$\times$ \\
Linux & algebra\_order & 5 & 2.301 & 2.317 & 0.046 & 0.043 & 2.516 & 78.2$\times$ \\
Linux & sets\_functions & 5 & 2.233 & 2.257 & 0.011 & 0.009 & 2.383 & 245.8$\times$ \\
Linux & number\_theory\_nat & 5 & 2.199 & 2.217 & 0.009 & 0.008 & 2.390 & 322.0$\times$ \\
\midrule
macOS & tcs\_binary\_heap & 9 & 2.576 & 2.931 & 0.049 & 0.042 & 2.589 & 155.9$\times$ \\
macOS & tcs\_treap\_analysis & 2 & 2.082 & 2.219 & 0.034 & 0.034 & 2.181 & 77.8$\times$ \\
macOS & tcs\_weighted\_graph & 9 & 2.617 & 2.461 & 0.031 & 0.028 & 2.603 & 194.9$\times$ \\
Linux & tcs\_binary\_heap & 9 & 1.886 & 1.807 & 0.054 & 0.051 & 1.877 & 103.2$\times$ \\
Linux & tcs\_treap\_analysis & 2 & 1.495 & 1.441 & 0.036 & 0.040 & 1.560 & 53.1$\times$ \\
Linux & tcs\_weighted\_graph & 9 & 1.771 & 1.683 & 0.032 & 0.034 & 1.761 & 127.5$\times$ \\
\bottomrule
\end{tabular}
\end{table}

\begin{table}[H]
\centering
\caption{LeanProbe sequential same-file results. Each file has five declarations and ten partial/full scenarios. Times are total seconds for the file.}
\label{tab:appendix-leanprobe-sequential}
\footnotesize
\setlength{\tabcolsep}{3pt}
\begin{tabular}{llrrrrrrrrr}
\toprule
Platform & File & Decls. & Scen. & Lake prefix & Lake full & Probe cached & Probe fresh & vs prefix & vs full & vs fresh \\
\midrule
macOS & analysis\_real & 5 & 10 & 67.992 & 40.216 & 4.775 & 44.699 & 14.24$\times$ & 8.42$\times$ & 9.36$\times$ \\
macOS & algebra\_order & 5 & 10 & 46.870 & 48.163 & 4.339 & 40.953 & 10.80$\times$ & 11.10$\times$ & 9.44$\times$ \\
macOS & sets\_functions & 5 & 10 & 45.421 & 42.237 & 3.916 & 37.797 & 11.60$\times$ & 10.79$\times$ & 9.65$\times$ \\
macOS & number\_theory\_nat & 5 & 10 & 36.474 & 36.648 & 3.789 & 36.736 & 9.63$\times$ & 9.67$\times$ & 9.70$\times$ \\
Linux & analysis\_real & 5 & 10 & 22.765 & 22.958 & 2.515 & 24.890 & 9.05$\times$ & 9.13$\times$ & 9.90$\times$ \\
Linux & algebra\_order & 5 & 10 & 23.186 & 23.489 & 2.547 & 25.147 & 9.10$\times$ & 9.22$\times$ & 9.87$\times$ \\
Linux & sets\_functions & 5 & 10 & 22.679 & 22.567 & 2.384 & 24.195 & 9.51$\times$ & 9.47$\times$ & 10.15$\times$ \\
Linux & number\_theory\_nat & 5 & 10 & 22.593 & 22.829 & 2.301 & 24.086 & 9.82$\times$ & 9.92$\times$ & 10.47$\times$ \\
\bottomrule
\end{tabular}
\end{table}

Repeated-target rows isolate the cost of checking many replacements for one active declaration after a single prepare step.
For $n$ complete replacement attempts on the same target, the comparison is
\[
  \text{Probe total}(n)=\text{Prepare}+n\cdot\text{Cached},
  \qquad
  \text{Lake total}(n)=n\cdot\text{Lake full}.
\]
The preparation cost is therefore amortized as soon as a target receives several candidate repairs: after setup, each additional candidate costs the cached-check time instead of another full-file \texttt{lake env lean} run.
The grouped rows report this steady-state behavior directly through \emph{Cached}, \emph{Fresh}, and \emph{Fresh/cached}; the sequential rows report the end-to-end cost when successful declarations advance the environment.
Sequential rows model queue-style proving in one file: partial \texttt{sorry} scenarios are diagnosed but not cached, while complete accepted declarations advance the environment for later targets.
This matches the proof-loop policy used in our runs: keep the LeanProbe server warm, prepare once before a target when several repairs are likely, and advance the cached environment only after a complete declaration has been accepted.

\end{document}